\documentclass[letterpaper, 10 pt, conference]{ieeeconf}  

\IEEEoverridecommandlockouts                              

\usepackage{graphicx} 
\usepackage{epsfig} 
\usepackage{mathptmx} 
\usepackage{times} 
\usepackage{amsmath} 
\usepackage{amssymb}  
\usepackage{cite}
\usepackage{xcolor}     
\usepackage{caption}  
\usepackage{multirow}
\usepackage{lipsum}   
\usepackage{placeins} 
\usepackage{float} 
\usepackage{marvosym}
\usepackage{hyperref}
\title{\LARGE \bf
KAN We Flow? Advancing Robotic Manipulation with 3D Flow Matching via KAN \& RWKV
}

\author{Zhihao Chen\textsuperscript{1}, Yiyuan Ge\textsuperscript{2, \Letter}, Ziyang Wang\textsuperscript{3}, Youwei Zhang\textsuperscript{4}
\thanks{\textsuperscript{1} Zhihao Chen is with the School of Intelligent Engineering and Automation,
Beijing University of Posts and Telecommunications (BUPT), China,
and also with Beijing Hydrogen Intelligence Technology Co. Ltd., China.
        {\tt\small zhihaochen666@bupt.edu.cn}}%
\thanks{\textsuperscript{2, \Letter} Yiyuan Ge (corresponding author) is with the School of Electronic and Information Engineering, South China University of Technology, China. 
        {\tt\small 202510182871@mail.scut.edu.cn}}%
\thanks{\textsuperscript{3} Ziyang Wang is with the Department of Computer Science at the University of Oxford, UK.
        {\tt\small ziyangwang@ieee.org}}%
\thanks{\textsuperscript{4} Youwei Zhang is with the School of Computer Engineering and Science, Shanghai University, China.
        {\tt\small yw-zhang@shu.edu.cn}}
\thanks{This work was supported by the Young Scientists Fund of NSFC (Grant No. 62406035), and Start-up Funding of Beijing University of Posts and Telecommunications (grant No.510224072).}
}

\begin{document}

\maketitle
\thispagestyle{empty}
\pagestyle{empty}

\begin{abstract}

Diffusion-based visuomotor policies excel at modeling action distributions but are inference-inefficient, since recursively denoising from noise to policy requires many steps and heavy UNet backbones, which hinders deployment on resource-constrained robots. Flow matching alleviates the sampling burden by learning a one-step vector field, yet prior implementations still inherit large UNet-style architectures. In this work, we present KAN-We-Flow, a flow-matching policy that draws on recent advances in Receptance Weighted Key Value (RWKV) and Kolmogorov-Arnold Networks (KAN) from vision to build a lightweight and highly expressive backbone for 3D manipulation. Concretely, we introduce an RWKV-KAN block: an RWKV first performs efficient time/channel mixing to propagate task context, and a subsequent GroupKAN layer applies learnable spline-based, groupwise functional mappings to perform feature-wise nonlinear calibration of the action mapping on RWKV outputs. Moreover, we introduce an Action Consistency Regularization (ACR), a lightweight auxiliary loss that enforces alignment between predicted action trajectories and expert demonstrations via Euler extrapolation, providing additional supervision to stabilize training and improve policy precision. Without resorting to large UNets, our design reduces parameters by 86.8\%, maintains fast runtime, and achieves state-of-the-art success rates on Adroit, Meta-World, and DexArt benchmarks. Our project page can be viewed in \href{https://1024ailab.github.io/KAN-We-Flow.github.io/}{\textcolor{red}{link}}
.
\end{abstract}

\section{INTRODUCTION}
Imitation learning \cite{argall2009survey, wang2024diffail, wang2024rise, agarwal2023dexterous, haldar2023teach} enables robots to acquire and reproduce specific skills by observing expert demonstrations, allowing them to execute delicate, highly complex behaviors that are difficult to specify with conventional programming. Among current approaches, generative-model-based policies \cite{lu2024manicm, dp3, mambapolicy, diffusionpolicy, yang2025fp3, zhang2025flowpolicy, mp1, adaflow, postels2022maniflow, zhi20253dflowaction, yang2023large, cao2025image, cao2025controllable} have become dominant, and they transfer advances from text-to-image generation to robotic imitation learning, achieving strong performance.

These generative methods can be broadly classified into
two types: diffusion-based approaches \cite{lu2024manicm, dp3, mambapolicy, diffusionpolicy, yang2025fp3} and flow matching paradigms \cite{zhang2025flowpolicy, mp1, postels2022maniflow, zhi20253dflowaction, adaflow}. Diffusion-based approaches model multi-modal action distributions by iterative denoising, which has proven effective in robotic manipulation. However, a key limitation of diffusion policies is slow inference speed, producing each action requires many sampling steps, leading to high latency, unsuitable for real-time control \cite{zhang2025flowpolicy, mp1, adaflow}. To address this issue, flow matching methods are introduced, which directly learn a continuous vector field that transports noise to target actions, often enabling one-step generation of actions \cite{postels2022maniflow, zhi20253dflowaction}. Despite the strong success rates of both diffusion and flow-matching policies (Fig.~\ref{fig:example} a), practical deployment remains bottlenecked by heavy UNet-style backbones that inflate parameters (Fig.~\ref{fig:example} b). In diffusion, DP3 \cite{dp3} is a representative case whose UNet exceeds $\sim$255 M parameters, and even a simplified DP3 speeds up mainly by trimming the UNet, underscoring the backbone as the dominant compute sink. For flow matching, they alleviate sampling inefficiency by directly learning a vector field, which often enables one-step action generation. It also delivers a runtime per step faster than diffusion baselines; for example, FlowPolicy \cite{zhang2025flowpolicy} runs around 20 ms in the Adroit Pen task and is about 7 times faster than DP3 \cite{dp3}. However, its flow implementations still inherit large UNet-like backbones (see Fig.~\ref{fig:example} b), which limit edge deployment. To make visuomotor policies practical on resource-constrained robots, we thus seek an architecture that preserves the accuracy and inference speed of flow models while substantially reducing compute and sustaining long-horizon prediction, which are crucial for robot learning. 

\begin{figure}[t] 
\centering 
\includegraphics[width=\linewidth]{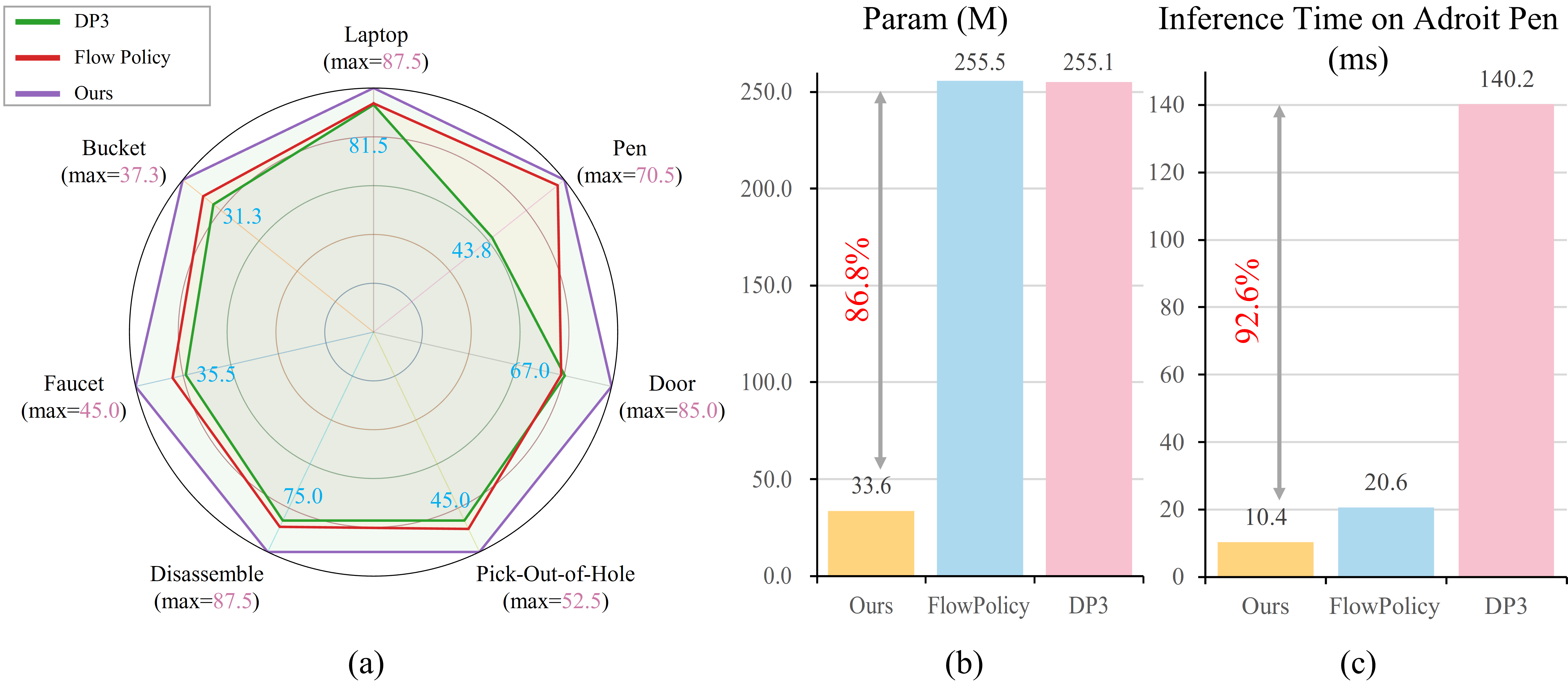} 
\caption{Comparison of KAN-We-Flow with the state-of-the-art methods FlowPolicy and DP3 regarding accuracy, parameter, and inference time. (a) KAN-We-Flow achieves superior success rates among different benchmarks' tough tasks; (b) Our approach obtains an 86.8\% parameter reduction, compared with FlowPolicy and DP3; (c) Compared with DP3, our KAN-We-Flow achieves 92.6\% inference time decrease in the Adroit–Pen task, enabling real-time control.} 
\label{fig:example} 
\end{figure}

Recently, two efficient modeling advances, Receptance Weighted Key Value (RWKV) and Kolmogorov–Arnold Networks (KAN), offer an attractive path forward. RWKV attains transformer-level sequence modeling with linear complexity and has shown strong results across visual tasks \cite{chen2025zig, duan2024vision, fei2024diffusion, zhou2024bsbp}. KAN, grounded in the Kolmogorov–Arnold representation theorem, replaces heavy matrix-based feature mixing with stacks of learnable 1D nonlinearities (spline-like activations), enabling efficient approximation of multivariate functions with far fewer parameters \cite{ukan, somvanshi2024survey, aghaei2024rkan}. Although these designs have demonstrated compelling efficiency in vision (e.g., Vision-RWKV \cite{duan2024vision} and U-KAN \cite{ukan}), their application in the robotic domain remains largely unexplored, which motivates us to explore their potential to achieve efficient manipulation.

In this paper, we aim to propose an efficient RWKV and KAN-based policy that reduces the large number of parameters in previous methods and maintains the FlowPolicy's fast inference speed. We achieve this by introducing an RWKV and a GroupKAN layer, called the RWKV-KAN block. Specifically, RWKV’s recurrent sequence modeling absorbs temporal correlations across the action trajectory, while the GroupKAN layer compactly represents the intricate visuomotor mapping. By marrying RWKV’s sequential reasoning with KAN’s compact expressiveness, our policy network achieves drastically reduced model size and faster inference without sacrificing precision. To further stabilize training without extra inference steps, we introduce an Action Consistency Regularization (ACR), a lightweight auxiliary objective that aligns the policy’s predicted action trajectory with expert demonstrations via a simple Euler extrapolation from intermediate states to the end of the horizon. This adds a global end-point anchor to the flow-matching objective, reducing drift and exposure-bias effects while incurring negligible compute and no extra inference steps. Finally, we evaluate our approach across the Adroit, Meta-World, and DexArt benchmarks, and show that the KAN-We-Flow policy achieves state-of-the-art success rates in vision-based imitation learning. Our contributions are as follows:
\begin{itemize}
    \item We introduce a novel flow-matching policy that integrates RWKV and GroupKAN modules. This is the first use of these architectures for robotic visuomotor policy learning, enabling faster inference speed with fewer parameters compared to previous methods.
    \item We propose the Action Consistency Regularization (ACR) that anchors predicted actions to demonstrations at the horizon, improving stability and precision at negligible cost.
    \item We conduct extensive experiments on Adroit, Meta-World, and DexArt benchmarks, demonstrating that our KAN-We-Flow policy achieves higher success rates than state-of-the-art baselines.
\end{itemize}

\section{Related Work}

\subsection{Diffusion vs. Flow-Based Policies}

Generative visuomotor policies for robotic manipulation mainly follow two paradigms: diffusion-based~\cite{diffusionpolicy, dp3, mambapolicy, hou2024diffusion, ma2024hierarchical, yang2025large, feng2025embodied, feng2025evoagent} and flow-matching–based~\cite{su2025freqpolicy, mp1, gkanatsios20253d, zhi20253dflowaction, zhang2025flowpolicy}. 
Diffusion policies model actions via conditional denoising, which is expressive and often stable to train, but typically require many iterative sampling steps and heavy backbones, leading to high latency and limited edge deployability (e.g., large UNet-style models in DP3~\cite{dp3}). 
Recent variants reduce backbone or improve long-horizon control (e.g., state-space/transformer backbones~\cite{mambapolicy, hou2024diffusion} and hierarchical designs~\cite{ma2024hierarchical}), yet the iterative sampling bottleneck remains.

Flow-matching methods instead generate actions in one or a few steps, enabling real-time inference and competitive success via temporal/frequency constraints and tailored regularization~\cite{su2025freqpolicy, mp1, gkanatsios20253d}. 
However, many implementations still depend on large UNet-like backbones or multi-stage pipelines (e.g., engineered refinement loops~\cite{gkanatsios20253d} or VLM-in-the-loop planning~\cite{zhi20253dflowaction}), which can reintroduce compute overhead.

Overall, diffusion offers strong expressivity but is inference-inefficient, while flow matching improves inference efficiency yet often inherits heavyweight backbones.
To address this, we propose a lightweight flow-matching policy that integrates RWKV and GroupKAN modules, achieving faster inference with fewer parameters than prior generative policies.

\begin{figure*}[!t]
\centering
\includegraphics[width=\textwidth]{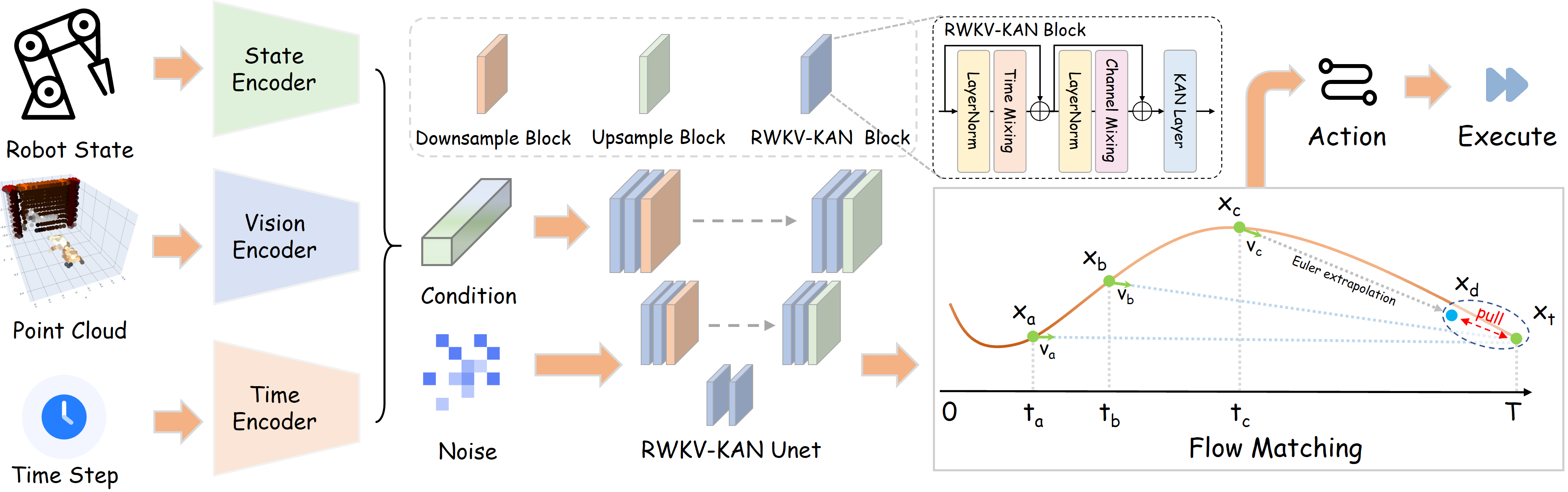} 
\caption{\textbf{Overview of KAN-We-Flow.} The policy receives a noised action and a condition that comprises three encoded parts, a point-cloud perception embedding, a robot-state embedding, and a time embedding. The concatenated representation is processed by a lightweight RWKV-KAN U-shaped backbone instead of a large UNet-style backbone; RWKV mixes long-range time/channel context with linear complexity, while KAN performs learnable spline-based feature calibration. Then, a straight-line flow is learned with conditional consistency flow matching to produce a one-step velocity field, and the resulting actions are generated at real-time inference speed; an additional action consistency regularization aligns Euler-extrapolated trajectories with demonstrations to stabilize training.}
\label{fig2}
\end{figure*}

\subsection{Current Policy Network Architectures}

Modern visuomotor policy backbones can be categorised into three principal paradigms: (i) convolutional neural networks \cite{dp3, zhang2025flowpolicy, mp1}, (ii) transformers \cite{dasari2024ingredients, hou2024diffusion, wang2025mtdp}, and (iii) emerging state-space (Mamba) models \cite{mambapolicy, ge2024mambatsr}.

For the CNN paradigm, DP3 \cite{dp3} marries compact 3D point-cloud encodings with a diffusion UNet to denoise noise into action sequences, delivering strong generalization across spatial/appearance shifts while remaining diffusion-style in inference. FlowPolicy \cite{zhang2025flowpolicy} keeps the CNN backbone but replaces iterative denoising with consistency flow matching so that a single network evaluation yields actions with ~7× faster inference yet competitive success. MeanFlow \cite{mp1} also uses point-cloud inputs with a lightweight CNN stack, learning interval-averaged velocities to achieve genuine 1-NFE trajectory generation and millisecond-level latency while outperforming DP3 and FlowPolicy. 

For transformer-based designs, DiT-Block \cite{dasari2024ingredients} Policy builds an encoder–decoder diffusion Transformer with adaLN-Zero stabilization, scaling to long-horizon dexterous tasks and multi-modal conditioning. Diffusion Transformer Policy \cite{hou2024diffusion} directly uses a large causal Transformer to denoise continuous action chunks from images/language, achieving strong generalist performance but still requiring multiple denoising steps at test time. MTDP \cite{wang2025mtdp} introduces a Modulated Attention decoder that better fuses guiding conditions and, with a DDIM variant, reduces sampling steps while maintaining performance.

Finally, for state-space (Mamba) backbones, MambaPolicy \cite{mambapolicy} replaces heavy UNets with selective state-space modules, reporting $>$80\% parameter reductions and notable FLOPs savings without sacrificing success on Adroit, Meta-World, and DexArt.

\section{Method}

\subsection{Overview}
As shown in Fig.~\ref{fig2}, we first encode a single-view point cloud and robot state into a compact condition. This condition is then fed into a lightweight RWKV-KAN UNet, which predicts a one-step velocity field. RWKV provides linear-time temporal mixing to propagate long-horizon context, while GroupKAN performs parameter-efficient per-channel functional calibration. Next, we adopt conditional consistency flow matching (CFM) for single-step action generation, and introduce an Action Consistency Regularization (ACR) term. The ACR anchors the one-step decoded action at the horizon ($t \to 1$) to the expert demonstration via Euler extrapolation, which provides additional supervision to stabilize training without extra inference steps.

\subsection{RWKV-KAN UNet}
The RWKV-KAN UNet is a three-stage encoder–decoder where each stage stacks RWKV-KAN blocks: RWKV mixes long-range temporal context with linear complexity, while GroupKAN performs per-channel functional calibration in a parameter-efficient manner.
\subsubsection{RWKV}
\label{subsubsec:rwkv}
We adopt the RWKV's time-mixing and channel-mixing operations as the core of our module. Let $x_t\in\mathbb{R}^C$ denote the token at time $t$, and let $S(\cdot)$ be a temporal shift operator. We form the shifted input $\tilde{x}_t=S(x_t)$ and compute linear projections:
\begin{equation}
r_t = W_r \tilde{x}_t,\qquad
k_t = W_k \tilde{x}_t,\qquad
v_t = W_v \tilde{x}_t
\end{equation}
with $W_r,W_k,W_v\in\mathbb{R}^{C\times C}$. Then, we aggregate values with per-channel exponential time-decay $w\in\mathbb{R}^{C}$ and a separate ``current-token'' term via $u\in\mathbb{R}^C$:
\begin{align}
\tilde{v}^{\rightarrow}_t
&=\frac{\sum_{i=1}^{t-1}\exp\!\bigl(- (t-1-i) w + k_i\bigr)\odot v_i \;+\; \exp\!\bigl(u+k_t\bigr)\odot v_t}
{\sum_{i=1}^{t-1}\exp\!\bigl(- (t-1-i) w + k_i\bigr) \;+\; \exp\!\bigl(u+k_t\bigr)}
\label{eq:wkv-forward}
\end{align}
To leverage full trajectory context as in our implementation, we apply Eq.~\ref{eq:wkv-forward} on the time-reversed sequence as well and combine the two scans:
\begin{equation}
\tilde{v}_t \;=\; \tilde{v}^{\rightarrow}_t \;+\; \tilde{v}^{\leftarrow}_t 
\end{equation}
The time--mixing output is then gated by the receptance and linearly projected:
\begin{equation}
\mathrm{TM}(x_t) \;=\; W_o\!\left(\sigma(r_t)\odot \tilde{v}_t\right), \qquad W_o\in\mathbb{R}^{C\times C}
\end{equation}
Complementarily, the channel--mixing branch applies a token-wise gated MLP with squared-ReLU nonlinearity:
\begin{align}
r'_t &= W'_r x_t,\quad k'_t = W'_k x_t,\\
\mathrm{CM}(x_t) &= \sigma(r'_t)\odot \bigl(W'_v \,[\mathrm{ReLU}(k'_t)]^{\!2}\bigr)
\end{align}
where $W'_r,W'_k,W'_v\in\mathbb{R}^{C\times C}$. Finally, we employ a pre-norm residual, and the output $z_t$ can be described as:
\begin{equation}
y_t = x_t \;+\; \mathrm{CM}\!\bigl(\mathrm{LN}_2(x_t)\bigr), \quad
z_t = y_t \;+\; \mathrm{TM}\!\bigl(\mathrm{LN}_1(y_t)\bigr)
\end{equation} 


\begin{figure}[t] 
\centering 
\includegraphics[width=\linewidth]{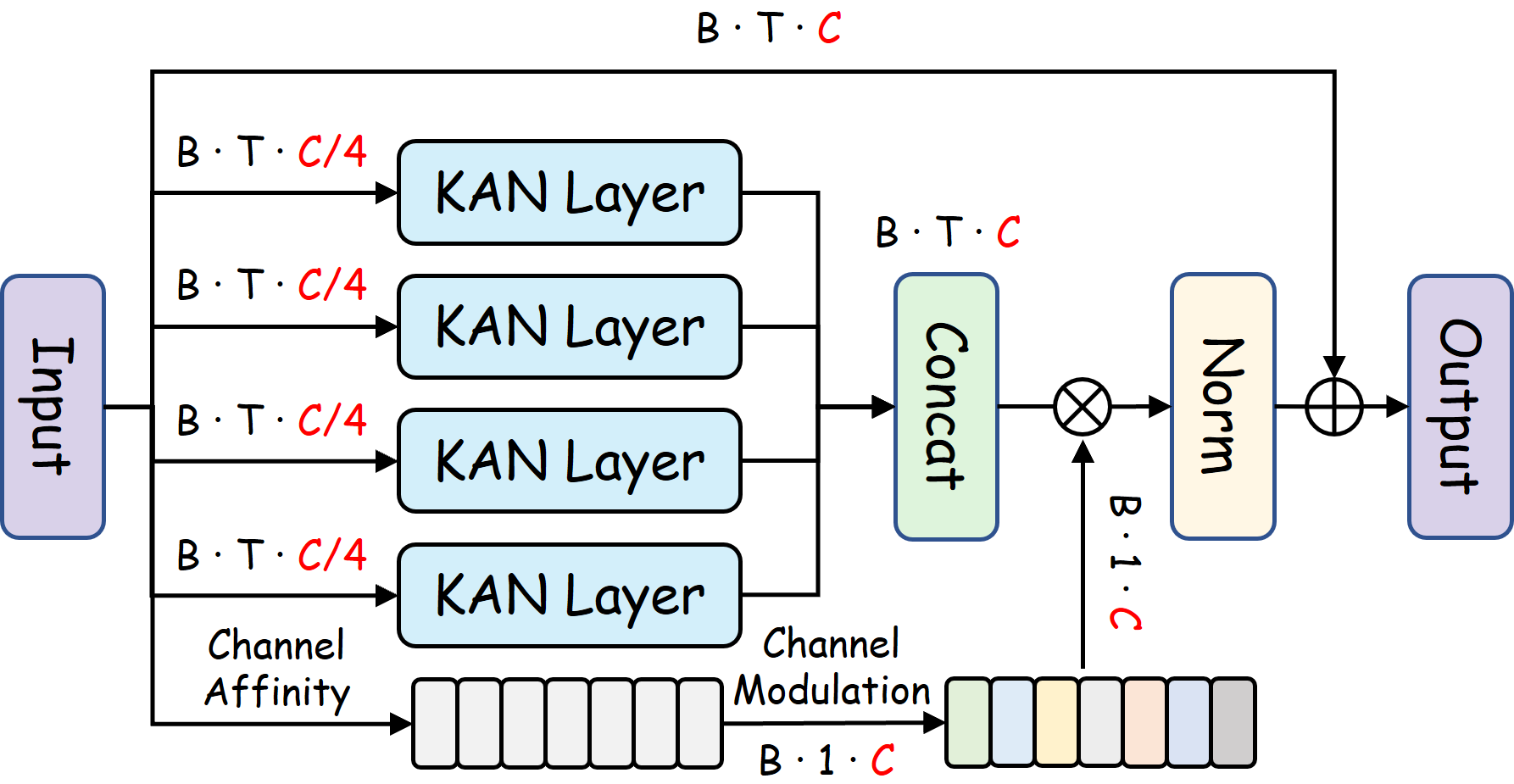} 
\caption{The architecture of the GroupKAN.} 
\label{fig:example} 
\end{figure}

\subsubsection{GroupKAN}
\label{subsec:GroupKAN}
We briefly recall the Kolmogorov–Arnold Network (KAN) formulation used in our design. A conventional $K$-layer MLP alternates linear maps and a fixed nonlinearity $\sigma$:
\begin{equation}
\mathrm{MLP}(\mathbf{Z}) \;=\; \big(W_{K-1}\circ\sigma\circ W_{K-2}\circ\sigma\circ\cdots\circ W_1\circ\sigma\circ W_0\big)\mathbf{Z}
\label{eq:mlp}
\end{equation}
whereas a $K$-layer KAN composes $K$ \emph{KAN layers} $\Phi_k$,
\begin{equation}
\mathrm{KAN}(\mathbf{Z}) \;=\; \big(\Phi_{K-1}\circ\Phi_{K-2}\circ\cdots\circ\Phi_{1}\circ\Phi_{0}\big)\mathbf{Z}
\label{eq:kan}
\end{equation}
each $\Phi_k$ acting from $\mathbb{R}^{n_k}\!\rightarrow\mathbb{R}^{n_{k+1}}$ via a \emph{matrix of learnable univariate functions} $\{\phi_{q,p}^{(k)}\}_{q,p}$ placed on edges rather than fixed node activations. Writing $\mathbf{Z}_{k+1}=\Phi_k\,\mathbf{Z}_{k}$, one has the matrix-of-functions form:
\begin{equation}
\Phi_k \;=\; 
\begin{pmatrix}
\phi^{(k)}_{1,1}(\,\cdot\,) & \phi^{(k)}_{1,2}(\,\cdot\,) & \cdots & \phi^{(k)}_{1,n_k}(\,\cdot\,)\\
\phi^{(k)}_{2,1}(\,\cdot\,) & \phi^{(k)}_{2,2}(\,\cdot\,) & \cdots & \phi^{(k)}_{2,n_k}(\,\cdot\,)\\
\vdots & \vdots & \ddots & \vdots\\
\phi^{(k)}_{n_{k+1},1}(\,\cdot\,) & \phi^{(k)}_{n_{k+1},2}(\,\cdot\,) & \cdots & \phi^{(k)}_{n_{k+1},n_k}(\,\cdot\,)
\end{pmatrix}\!
\label{eq:kan-matrix}
\end{equation}
which eliminates explicit weight matrices and instead learns spline-parameterized univariate functions on edges, a choice that improves parameter efficiency and interpretability while preserving approximation power.

Let $\mathbf{X}\in\mathbb{R}^{B\times T\times C}$ denote a sequence of $T$ time steps with $C$ channels. We partition the channel into $G$ equal groups, $\mathbf{X}=\mathrm{Concat}\big(\mathbf{X}_{1},\ldots,\mathbf{X}_{G}\big)$ with $\mathbf{X}_{g}\in\mathbb{R}^{B\times T\times C/G}$, and process each group by an \emph{independent} KAN operator that shares parameters across time. Concretely, flattening the first two axes, we define the operation process as:
\begin{equation}
\mathbf{Y}_{g} = \mathrm{KAN}_{g}(\mathbf{X}_{g})\in\mathbb{R}^{B\times T\times C/G},\quad g=1,\dots,G
\label{eq:groupkan-core}
\end{equation}
and concatenate the outputs along channels:
\begin{equation}
\mathbf{Y}\;=\;\mathrm{Concat}\big(\mathbf{Y}_{1},\ldots,\mathbf{Y}_{G}\big)\in\mathbb{R}^{B\times T\times C}
\label{eq:groupkan-concat}
\end{equation}
To adaptively reweight channels with sequence-aware statistics, we introduce a lightweight \emph{channel affinity modulation} (CAM). Let $\bar{\mathbf{X}}=\frac{1}{T}\sum_{t=1}^{T}\mathbf{X}_{[:,t,:]}\in\mathbb{R}^{B\times C}$ denote the temporal pooling outputs; CAM produces a gating vector $a \in R^{B \times C}$ and we broadcast it to $\mathbf{A}\in R^{B\times T\times C}$:

\begin{equation}
a=\sigma\!\Big(W_{2}\,\varphi\!\big(W_{1}\,\bar{\mathbf{X}}\big)\Big)\in\mathbb{R}^{B\times C}
\label{eq:cam}
\end{equation}
where $W_{1},W_{2}$ are learned linear maps, $\varphi$ is a smooth activation (e.g., SiLU), and $\sigma$ is the logistic sigmoid. The final GroupKAN's output can be denoted as:
\begin{equation}
\widehat{\mathbf{X}} = \mathbf{X} + \mathrm{DropPath}(\mathrm{LN}(\mathbf{A}\odot \mathbf{Y} ))
\label{eq:groupkan-block}
\end{equation}
where $\odot$ is elementwise multiplication. In practice, we use $G{=}4$ and each $\mathrm{KAN}_g$ acts on $C/4$ channels. GroupKAN preserves KAN’s efficient function approximation and integrates a CAM to highlight task-relevant channels, achieving parameter efficiency without compromising performance.


\subsection{Consistency Flow Matching Objective}
\label{sec:cfm}
We learn a velocity-consistent straight-line flow in the \emph{action} space, conditioned on robot state $s$ and visual representation $v$.
Let $a_{\text{src}}\!\sim\!\Gamma_{\text{src}} := \mathcal{N}(0,I)$ and $a_{\text{tar}}\!\sim\!\Gamma_{\text{tar}}(\cdot\,|\,s,v)$, where $a_{\text{src}}$ and $a_{\text{tar}}$ are action trajectories sampled from the source and target distributions, respectively; $\Gamma_{\text{src}}$ is the standard normal over the action space and $\Gamma_{\text{tar}}$ is the task distribution conditioned on $(s,v)$.
For any $t\in[0,1]$, define the linear interpolation between these samples:
\begin{equation}
\label{eq:interp}
a_t = (1-t)\,a_{\text{src}} + t\,a_{\text{tar}} \, 
\end{equation}
A time-conditioned vector field $v_\theta(a_t,t,s,v)$ induces a single-step decoder that advances from $t$ to $1$ via explicit Euler with remaining step size $(1-t)$:
\begin{equation}
\label{eq:decode}
f_\theta(t,a_t,s,v) = a_t + (1-t)\,v_\theta(a_t,t,s,v)\, 
\end{equation}
Here, $\theta$ denotes all learnable parameters of the velocity (flow) network $v_\theta$, and $f_\theta$ depends on $\theta$ only through $v_\theta$.


\paragraph{Consistency objective}
Consistency Flow Matching (CFM) enforces that decoding from two nearby times $t$ and $t{+}\Delta t$ yields the same endpoint, and that their instantaneous velocities agree. With an exponential moving average (EMA) and a small $\Delta t{>}0$, we minimize:
\begin{equation}
\label{eq:cfm_endpoint}
\mathcal L_{\text{end}}
=
\mathbb E_{t\sim\mu}\,\mathbb E_{a_t,\,a_{t+\Delta t}}
\Big[
\big\| f_\theta(t,a_t,s,v)-f_{\theta^-}(t{+}\Delta t,a_{t+\Delta t},s,v)\big\|_2^2
\Big]
\end{equation}
\begin{equation}
\label{eq:cfm_velocity}
\mathcal L_{\text{vel}}
=
\mathbb E_{t\sim\mu}\,\mathbb E_{a_t,\,a_{t+\Delta t}}
\Big[
\big\| v_\theta(a_t,t,s,v)-v_{\theta^-}(a_{t+\Delta t},t{+}\Delta t,s,v)\big\|_2^2
\Big]
\end{equation}
\begin{equation}
\label{eq:cfm}
\mathcal L_{\text{CFM}}
=
\mathcal L_{\text{end}}
+\alpha\,\mathcal L_{\text{vel}}
\end{equation}
where $\mu$ is uniform on $[0,1{-}\Delta t]$, $\mathbb{\alpha}$ denotes positive scalar, $\Delta t$ represents the time interval, and $\theta^-$ is the running average of past $\theta$ values. 

\paragraph{Multi-segment extension}
To relax global consistency and improve expressivity, we partition $[0,1]$ into $K$ uniform segments.
For segment $i\in\{0,\dots,K{-}1\}$, we optimize the segment-wise consistency:
\begin{equation}
\label{eq:seg_decode}
f^{(i)}_\theta(t,a_t,s,v)
\;=\;
a_t + \big((i{+}1)/K - t\big)\,v^{(i)}_\theta(a_t,t,s,v)
\end{equation}
\begin{equation}
\label{eq:seg_a}
a_i(t) \;=\; 
\big\| f^{(i)}_\theta(t,a_t,s,v)
      - f^{(i)}_{\theta^-}(t{+}\Delta t,\,a_{t+\Delta t},\,s,\,v)\big\|_2^2
\end{equation}
\begin{equation}
\label{eq:seg_b}
b_i(t) \;=\; 
\big\| v^{(i)}_\theta(a_t,t,s,v)
      - v^{(i)}_{\theta^-}(a_{t+\Delta t},\,t{+}\Delta t,\,s,\,v)\big\|_2^2
\end{equation}
\begin{equation}
\label{eq:seg_loss}
\mathcal L_{\text{MFM}}
\;=\;
\sum_{i=0}^{K-1}\lambda_i\,
\mathbb E_{t\sim\mu_i}\,\mathbb E_{a_t,\,a_{t+\Delta t}}
\big[\, a_i(t) + \alpha\, b_i(t) \,\big]
\end{equation}
where $\mu_i$ is uniform on $\big[i/K,(i{+}1)/K{-}\Delta t\big]$ and $\{\lambda_i\}$ balance segment importance.
In practice we use $K{=}2$ for a strong efficiency–quality trade-off.


\subsection{Action Consistency Regularization (ACR)}
\label{subsec:acr}
To explicitly anchor the \emph{one-step} decode at the terminal time, we introduce an action consistency regularizer that ties the one-step decode $f_\theta(t,a_t,s,v)$ at $t\!\to\!1$ to expert actions over a task-relevant control window.

Given the interpolation sample $a_t$ and conditions $c{:=}(s,v)$, we form a single-step decode using Eq.~\ref{eq:decode}:
\begin{equation}
\label{eq:acr_step}
\hat{a}_{1} \;=\; f_\theta(t,a_t,s,v) \;=\; a_t + (1-t)\,v_\theta(a_t,t,s,v)\, 
\end{equation}
Let the control window $W=\{u_0,\ldots,u_0{+}H{-}1\}$ denote the action horizon used for behavior matching, with $u_0=n_{\text{obs}}{-}1$ and $H$ the horizon length. 
Let $\mathcal{S}_W$ be the selector that extracts indices $W$ from a full action trajectory. 
We penalize the mean-squared deviation between the decoded actions and expert actions on $W$ (available \emph{only during training}):
\begin{equation}
\label{eq:acr_loss}
\mathcal{L}_{\mathrm{ACR}}
\;=\;
\frac{1}{|W|}\sum_{u\in W}\Bigl\|\bigl(\mathcal{S}_W \hat{a}_{1}\bigr)_u - a^{\star}_u\Bigr\|_2^2
\end{equation}
Intuitively, Eq.~\ref{eq:acr_step} performs a single-step decode from $(a_t,t)$ to $t{=}1$ via Eq.\ref{eq:decode}; Eq.~\ref{eq:acr_loss} enforces behavioral consistency with expert actions over $W$, biasing the learned vector field toward action-faithful solutions without extra network evaluations.

\subsection{Overall Objective and Inference}
Our final training loss combines the flow matching objective and the ACR term:
\begin{equation}
\label{eq:total}
\mathcal{L} \;=\; \mathcal{L}_{\text{MFM}} \;+\; \lambda_{ACR}\,\mathcal{L}_{ACR}
\end{equation}
where $\lambda_{\text{ACR}}\!\ge\!0$ (set to $1$ in our experiments). At test time, we draw $a_0\!\sim\!\mathcal{N}(0,I)$ and perform the one-step decode $a_1 = \Pi_a\!\big(x_t + (1-t)\,v_\theta(x_t,t,c)\big)$ with $t\!\in\!(0,1)$, yielding real-time action sequences while preserving high task success rates due to the ACR constraint.

\section{Experiment}
\subsection{Datasets}
We evaluate KAN-We-Flow on three benchmarks to verify its effectiveness, including 3 tasks in Adroit \cite{adroit}, 4 tasks in Dexart \cite{dexart}, and 34 tasks in Meta-World \cite{metaworld}. Following previous work \cite{dp3, zhang2025flowpolicy, mp1}, we adopt the same data collection method.

\subsection{Baselines}
To demonstrate the effectiveness of KAN-We-Flow, we compare it against both diffusion-based and flow-matching methods. For diffusion-based baselines, we include 3D diffusion policies (DP3 \cite{dp3}, Simple-DP3 \cite{dp3}, and MambaPolicy \cite{mambapolicy}) and 2D diffusion policies (DP \cite{diffusionpolicy}, BCRNN \cite{mandlekar2021matters}, and IBC \cite{florence2022implicit}). For flow-matching baselines, we consider Adaflow \cite{hu2024adaflow}, FlowPolicy \cite{zhang2025flowpolicy}, and MP1 \cite{mp1}. Because the expert demonstrations are generated stochastically, we re-implemented DP3 \cite{dp3}, MambaPolicy \cite{mambapolicy}, and FlowPolicy \cite{zhang2025flowpolicy} under the same environment and expert data as our method; we denote these runs with \dag\ to indicate our reimplementations and ensure a fair comparison.

\begin{figure}[!t]
\centering
\includegraphics[width=\columnwidth]{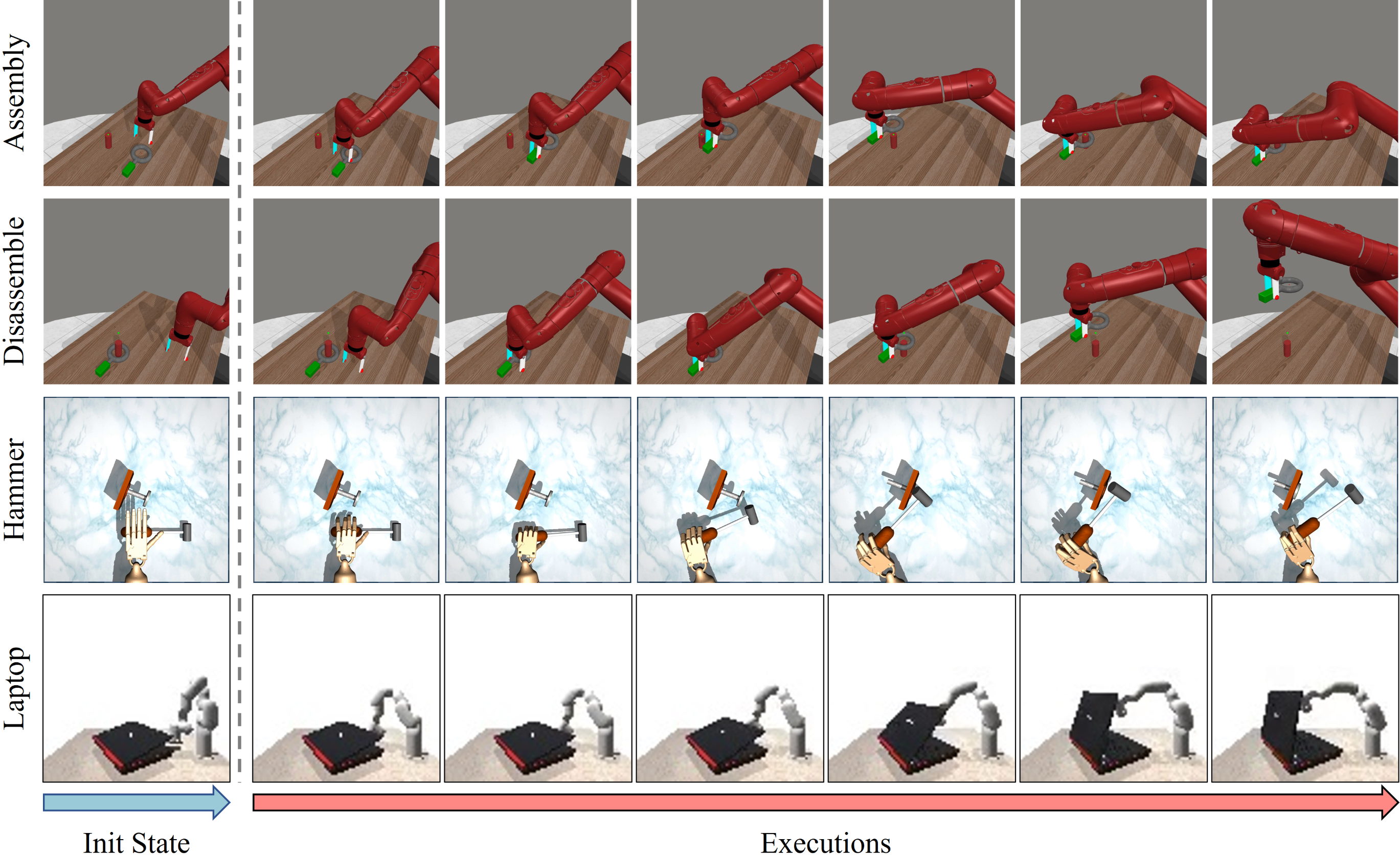} 
\caption{Visualization of manipulation results. We evaluate on three datasets: Meta-World, Adroit, and DexArt. We illustrate representative rollouts from Meta-World (Assembly, Disassemble), Adroit (Hammer), and DexArt (Laptop). During interaction, the KAN-We-Flow policy predicts future action sequences and continues issuing actions until the task is successfully completed.}
\label{fig:fig2}
\end{figure}

\begin{table*}[!t]
\centering
\caption{Quantitative comparison of baselines in simulated environments. We compare KAN-We-Flow against diffusion-based and flow-matching methods on the Adroit, Meta-World, and DexArt datasets. \dag\ indicates results reproduced by us for fair comparison. KAN-We-Flow achieves the best performance across all domains.}
\label{tab:table1}
\setlength{\tabcolsep}{3pt} 
\scriptsize                 
\resizebox{\linewidth}{!}{%
\begin{tabular}{c|ccc|cccc|cccc}
\hline \hline
\multirow{2}{*}{Method} & \multicolumn{3}{c|}{Adroit} & \multicolumn{4}{c|}{DexArt} & \multicolumn{4}{c}{Meta-World} \\
& Hammer & Door & Pen & Laptop & Faucet & Toilet & Bucket & Easy (21) & Medium (4) & Hard (4) & Very Hard (5) \\ \hline
Diffusion Policy \cite{diffusionpolicy} & 48.0$\pm$17.0 & 50.0$\pm$5.0 & 25.0$\pm$4.0 & 69.0$\pm$4.0 & 23.0$\pm$8.0 & 58.0$\pm$2.0 & 46.0$\pm$1.0 & 50.7$\pm$6.1 & 11.0$\pm$2.5 & 5.25$\pm$2.5 & 22.0$\pm$5.0 \\
BCRNN \cite{mandlekar2021matters} & 8.0$\pm$14.0 & 0.0$\pm$0.0 & 8.0$\pm$1.0 & 29.0$\pm$12.0 & 26.0$\pm$2.0 & 38.0$\pm$10.0 & 24.0$\pm$11.0 & - & - & - & - \\
IBC \cite{florence2022implicit} & 0.0$\pm$0.0 & 0.0$\pm$0.0 & 10.0$\pm$1.0 & 1.0$\pm$1.0 & 7.0$\pm$2.0 & 15.0$\pm$1.0 & 0.0$\pm$0.0 & - & - & - & - \\
DP3 \cite{dp3} \dag \ & 100.0$\pm$0.0 & 65.0$\pm$3.3 & 42.3$\pm$1.5 & 78.0$\pm$3.5 & 32.0$\pm$3.5 & 75.3$\pm$2.5 & 29.0$\pm$2.3 & 88.3$\pm$3.2 & 45.5$\pm$8.5 & 33.7$\pm$8.7 & 42.0$\pm$9.0 \\
Simple DP3 \cite{dp3} \dag \ & 98.0$\pm$2.0 & 50.0$\pm$2.3 & 40.0$\pm$1.3 & 75.0$\pm$2.3 & 30$\pm$2.0 & 70.0$\pm$2.0 & 25.0$\pm$1.3 & 87.0$\pm$2.0 & 43.0$\pm$7.0 & 30.0$\pm$6.0 & 36.0$\pm$6.0 \\
MambaPolicy \cite{mambapolicy} \dag \ & 100.0$\pm$0.0 & 70.0$\pm$1.0 & 42.0$\pm$2.3 & 80.0$\pm$1.0 & 35.0$\pm$2.0 & 77.0$\pm$1.0 & 28.0$\pm$1.0 & 89.0$\pm$1.0 & 46.0$\pm$7.5 & 34.3$\pm$9.3 & 43.0$\pm$8.0 \\
Adaflow \cite{adaflow} & 45.0$\pm$11.0 & 27.0$\pm$6.0 & 18.0$\pm$6.0 & - & - & - & - & 50.4$\pm$7.0 & 14.0$\pm$6.0 & 8.0$\pm$5.0 & 25.0$\pm$6.0 \\
FlowPolicy \cite{zhang2025flowpolicy} \dag \ & 100.0$\pm$0.0 & 62.0$\pm$5.0 & 55.0$\pm$13.0 & 81.0$\pm$1.0 & 34.0$\pm$4.0 & 77.0$\pm$3.0 & 30.0$\pm$3.3 & 85.0$\pm$3.3 & 59.0$\pm$8.0 & 42.0$\pm$5.0 & 53.0$\pm$6.0 \\
MP1 \cite{mp1} & 100.0$\pm$0.0 & 69.0$\pm$2.0 & 58.0$\pm$5.0 & - & - & - & - & 88.2$\pm$1.1 & 68.0$\pm$3.1 & 58.1$\pm$5.0 & 67.2$\pm$2.7 \\ \hline
\textbf{Ours} & \textbf{100.0$\pm$0.0} & \textbf{83.0$\pm$2.0} & \textbf{68.0$\pm$2.5} & \textbf{85.0$\pm$2.5} & \textbf{42.5$\pm$2.5} & \textbf{82.0$\pm$2.0} & \textbf{35.0$\pm$2.3} & \textbf{92.0$\pm$1.0} & \textbf{72.0$\pm$1.3} & \textbf{63.3$\pm$1.3} & \textbf{71.3$\pm$1.0} \\ \hline \hline
\end{tabular}%
}
\end{table*}

\subsection{Implementation Details}
We generate ten expert demonstrations for training on Adroit \cite{adroit}, Dexart \cite{dexart}, and Meta-World \cite{metaworld}. Like the previous works \cite{mambapolicy, dp3}, the image size is cropped to a resolution of $84 \times 84$ pixels, and the point cloud is downsampled to $512$ points for Adroit or $1024$ points for Meta-World using farthest point sampling (FPS). We set the prediction horizon to $4$, the observation length to $2$, and the action prediction length to $3$. We adopt an AdamW optimizer to update the weights during the training phase, with a learning rate of $1e-4$ and a batch size of $128$. The model is trained for 3,000 epochs. We use an EMA decay of 0.95. During training, both actions and states are normalized to the range [-1, 1]; actions are de-normalized before execution. All implementations, including ours and the baselines, are built in PyTorch and evaluated on a single NVIDIA RTX 3090 GPU.

\subsection{Evaluation Metrics}
We report top-1, top-3, and top-5 success rates ($SR1$, $SR3$, $SR5$), computed as the average of the highest $1$, $3$, and $5$ trial outcomes per task, respectively. For each domain, we use three random seeds $(0, 42, 100)$ and report the mean and standard deviation across the seeds.

\begin{table}[!t]
\centering
\caption{Comparison of inference time across methods on the Meta-World benchmark. Because of the multi-step denoising procedure, diffusion-based methods are slower than flow-based methods. KAN-We-Flow achieves state-of-the-art inference speed across all subtasks.}
\label{tab:table2}
\resizebox{\columnwidth}{!}{
\begin{tabular}{c|ccc|cccc}
\hline \hline
\multirow{2}{*}{Method} & \multicolumn{3}{c|}{Adroit/ms}                                                             & \multicolumn{4}{c}{Meta-World/ms}                                                                                         \\
                        & Hammer                       & Door                         & Pen                          & Easy (21)                    & Medium (4)                   & Hard (4)                     & Very Hard (5)               \\ \hline
DP3 \cite{dp3}                    & 129.5$\pm$13.9 & 141.3$\pm$14.0 & 130.0$\pm$10.2 & 130.0$\pm$10.5 & 132.0$\pm$10.3 & 130.0$\pm$10.5 & 135.0$\pm$10.5 \\
Simple DP3 \cite{dp3}              & 103.1$\pm$11.4 & 111.3$\pm$10.2 & 115.0$\pm$3.0  & 91.9$\pm$8.6   & 98.3$\pm$9.1   & 101.3$\pm$9.7  & 103.8$\pm$10.2 \\
FlowPolicy \cite{zhang2025flowpolicy}             & 15.3$\pm$1.1   & 13.2$\pm$4.0   & 20.0$\pm$0.6   & 12.0$\pm$1.4   & 12.2$\pm$1.5   & 13.5$\pm$1.4   & 14.5$\pm$1.6   \\
\textbf{Ours}                    & \textbf{9.5$\pm$2.3}    & \textbf{8.0$\pm$2.5}    & \textbf{10.1$\pm$0.3}   & \textbf{10.1$\pm$1.0}   & \textbf{6.7$\pm$0.1}    & \textbf{10.3$\pm$1.1}   & \textbf{11.8$\pm$1.1}   \\ \hline \hline
\end{tabular}
}
\end{table}

\subsection{Comparisons with the State-of-the-Art}
\begin{figure}[t] 
\centering 
\includegraphics[width=\linewidth]{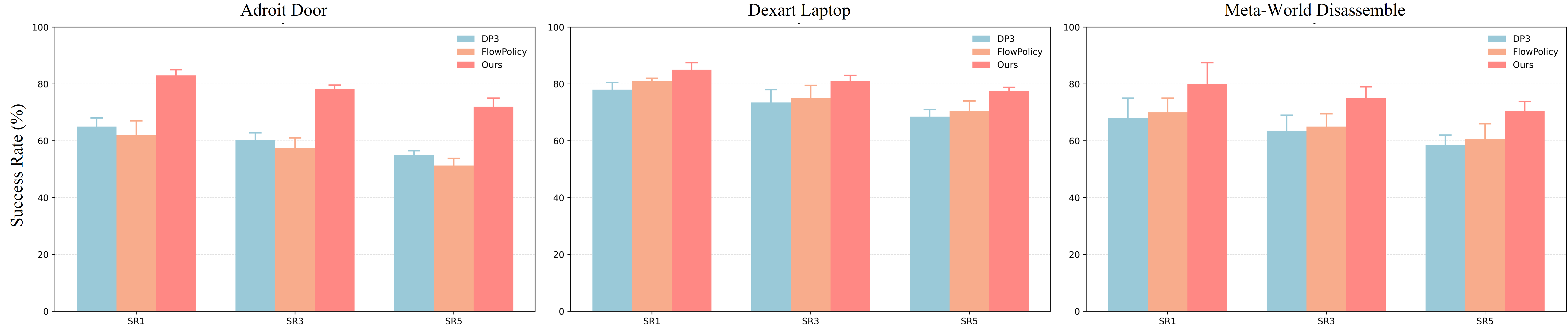} 
\caption{Visualization of success rates. We plot top-1, top-3, and top-5 success rates (SR1, SR3, SR5), computed as the average of the highest 1, 3, and 5 trial outcomes per task; KAN-We-Flow achieves superior performance across tasks.} 
\label{fig:fig4} 
\end{figure}

We compare KAN-We-Flow against both diffusion-based (DP3, Simple-DP3, MambaPolicy, Diffusion Policy, BCRNN, IBC) and flow-based (Adaflow, FlowPolicy, MP1) policies on Adroit, DexArt, and Meta-World. As summarized in Table~\ref{tab:table1}, our method achieves the best overall success rates across all domains and difficulty tiers. On \emph{Adroit}, KAN-We-Flow raises Door and Pen to 83.0$\pm$2.0 and 68.0$\pm$2.5, outperforming FlowPolicy (62.0$\pm$5.0, 55.0$\pm$13.0) and DP3 (65.0$\pm$3.3, 42.3$\pm$1.5). On \emph{DexArt}, it improves over FlowPolicy across Faucet, Toilet, Bucket, and Laptop. On \emph{Meta-World}, it exceeds both MP1 and FlowPolicy across \emph{Easy/Medium/Hard/Very Hard}; for example, 92.0$\pm$1.0 vs. 88.2$\pm$1.1 (MP1) and 85.0$\pm$3.3 (FlowPolicy) on Easy, and 71.3$\pm$1.0 vs. 67.2$\pm$2.7 (MP1) and 53.0$\pm$6.0 (FlowPolicy) on Very Hard. These aggregate gains are visualized in Fig.~\ref{fig:fig4} through $SR1/SR3/SR5$ bars, highlighting consistent improvements not only at top-1 but also at the top-$K$ operating points used in prior work. 

\subsection{Discussion}
We attribute the accuracy gains to two complementary design choices: (i) the \emph{RWKV–KAN} UNet, which captures long-horizon temporal dependencies and per-channel functional calibration with far fewer parameters than previous UNet-style stacks; and (ii) the \emph{Action Consistency Regularization (ACR)}, which anchors the one-step decode to expert behaviors and reduces drift, particularly beneficial on longer-horizon, harder tasks. Overall, these results validate the central thesis from the introduction: replacing heavy UNet-style backbones with RWKV–KAN UNet and adding a lightweight ACR term yields a policy that is both more \emph{accurate} and \emph{deployable} across diverse manipulation benchmarks.

\subsection{Efficiency Analysis}
Beyond accuracy, KAN-We-Flow is markedly more efficient in parameters and runtime. Table~\ref{tab:table3} shows that our model uses only \textbf{33.6M} parameters and \textbf{0.03G} FLOPs with \textbf{101.2 MB} GPU memory, cutting parameters by \textbf{86.8\%} vs.\ DP3 (255.1M) and still improving success on challenging Meta-World tasks such as Stick-Pull (VH), Disassemble (VH), and Pick-Out-Of-Hole (H). Inference speed mirrors these savings: as reported in Table~\ref{tab:table3}, KAN-We-Flow runs in \textbf{8.0–10.1 ms} on Adroit Door/Pen/Hammer and \textbf{6.7–11.8 ms} across Meta-World tiers, comfortably supporting $\sim$100 Hz control. This is substantially faster than diffusion baselines (e.g., DP3 $\sim$103–141 ms) and improves over FlowPolicy’s already fast flow-matching inference (e.g., Door 13.2$\pm$4.0 ms $\rightarrow$ \textbf{8.0}$\pm$2.5 ms). The combination of a one-step CFM decoder and the RWKV–KAN UNet therefore delivers both latency and memory wins without sacrificing accuracy. In short, KAN-We-Flow meets practical needs by enabling real-time control with compact models that fit edge compute budgets. It also outperforms prior SOTA methods, achieving precisely the efficiency–accuracy balance we targeted.

\begin{table*}[!t]
\centering
\caption{We compare parameter efficiency across DP3, FlowPolicy, MambaPolicy, and KAN-We-Flow. In the challenging Meta-World settings, our method achieves stronger results with 86.8\% fewer parameters compared with DP3. We abbreviate the Very Hard and Hard difficulty levels as VH and H.}
\label{tab:table3}
\resizebox{\linewidth}{!}{
\begin{tabular}{c|ccc|ccc}
\hline \hline
\multirow{2}{*}{Method} & \multirow{2}{*}{Param (M)} & \multirow{2}{*}{FLOPs (G)} & \multirow{2}{*}{GPU Usage (MB)} & \multicolumn{3}{c}{Hard Environment in Meta-World} \\
                        &                            &                            &                                 & Stick-Pull (VH)   & Disassemble (VH)    & Pick-Out-Of-Hole (H)    \\ \hline
DP3 \cite{dp3}                    & 255.1                      & 0.3                        & 996.1                           & 82.6          & 75.0           & 45.0                \\
FlowPolicy \cite{zhang2025flowpolicy}             & 255.5                      & 0.4                        & 980.2                           & 83.4          & 78.5           & 47.5                \\
MambaPolicy \cite{mambapolicy}            & 47.9                       & 0.03                       & 137.7                           & 86.7          & 81.7           & 48.3                \\
\textbf{Ours}             & \textbf{33.6}                       & \textbf{0.03}                       & \textbf{101.2}                           & \textbf{90.2}          & \textbf{87.5}           & \textbf{52.5}                \\ \hline \hline
\end{tabular}
}
\end{table*}

\begin{table}[t]
\centering
\caption{Ablation study of key components of KAN-We-Flow in the Adroit Door environment. The results highlight the necessity of each proposed component.}
\label{tab:table4}
\resizebox{\columnwidth}{!}{
\begin{tabular}{ccccc|ccc}
\hline \hline
\multirow{2}{*}{Baseline} & \multirow{2}{*}{\begin{tabular}[c]{@{}c@{}}RWKV \\ Time-Mixing\end{tabular}} & \multirow{2}{*}{\begin{tabular}[c]{@{}c@{}}RWKV \\ Channel-Mixing\end{tabular}} & \multirow{2}{*}{GroupKAN} & \multirow{2}{*}{ACR} & \multicolumn{3}{c}{Adroit-Pen}                                                       \\
                          &                                   &                                      &                           &                      & $SR1$                        & $SR3$                        & $SR5$                        \\ \hline
$\checkmark$                         &                                   &                                      &                           &                      & 59.5$\pm$1.5 & 51.5$\pm$1.3 & 45.5$\pm$2.5 \\
$\checkmark$                         & $\checkmark$                                 &                                      &                           &                      & 62.0$\pm$1.3 & 52.5$\pm$2.3 & 48.3$\pm$1.3 \\
$\checkmark$                         & $\checkmark$                                 & $\checkmark$                                    &                           &                      & 64.3$\pm$2.0 & 55.5$\pm$2.0 & 50.5$\pm$0.3 \\
$\checkmark$                         & $\checkmark$                                 &                                      & $\checkmark$                         &                      & 65.0$\pm$3.3 & 58.3$\pm$1.3 & 53.3$\pm$2.3 \\
$\checkmark$                         &                                   & $\checkmark$                                    & $\checkmark$                         &                      & 65.5$\pm$2.3 & 58.0$\pm$1.5 & 54.0$\pm$2.0 \\
$\checkmark$                         & $\checkmark$                                 & $\checkmark$                                    & $\checkmark$                         & $\checkmark$                    & \textbf{68.0$\pm$2.5} & \textbf{61.3$\pm$1.0} & \textbf{56.5$\pm$1.5} \\ \hline \hline
\end{tabular}
}
\end{table}

\begin{table}[t]
\centering
\caption{Ablation of architectural components within the RWKV–KAN block on Adroit–Door. We evaluate replacements of RWKV with CNN/MHSA/Mamba-V1/V2 and GroupKAN with KAN/MLP. “r/w” denotes “replaced with”. }
\label{tab:door}
\resizebox{\columnwidth}{!}{
\begin{tabular}{l|ccc}
\hline \hline
\multicolumn{1}{c|}{\multirow{2}{*}{Method}} & \multicolumn{3}{c}{Adroit-Door}                                                      \\
\multicolumn{1}{c|}{}                        & $SR1$                        & $SR3$                        & $SR5$                        \\ \hline
w/o RWKV, r/w CNN block                      & 66.5$\pm$2.5 & 59.3$\pm$3.0 & 55.0$\pm$3.0 \\
w/o RWKV, r/w MHSA block \cite{dosovitskiy2020image}                    & 68.0$\pm$5.0 & 61.3$\pm$2.0 & 58.3$\pm$2.3 \\
w/o RWKV, r/w Mamba-V1 \cite{gu2023mamba}                       & 70.5$\pm$1.5 & 65.5$\pm$3.5 & 62.3$\pm$2.0 \\
w/o RWKV, r/w Mamba-V2 \cite{gu2023mamba}                      & 71.3$\pm$1.0 & 67.5$\pm$2.0 & 64.3$\pm$1.0 \\
w/o GroupKAN, r/w KAN \cite{ukan}                        & 81.0$\pm$0.5 & 76.0$\pm$0.5 & 70.5$\pm$0.5 \\
w/o GroupKAN, r/w MLP                        & 80.3$\pm$1.3 & 75.0$\pm$1.0 & 69.5$\pm$1.5 \\
\textbf{Ours}                                         & \textbf{83.0$\pm$2.0} & \textbf{78.3$\pm$1.3} & \textbf{72.0$\pm$3.0} \\ \hline \hline
\end{tabular}
}
\end{table}

\begin{figure}[t] 
\centering 
\includegraphics[width=\linewidth]{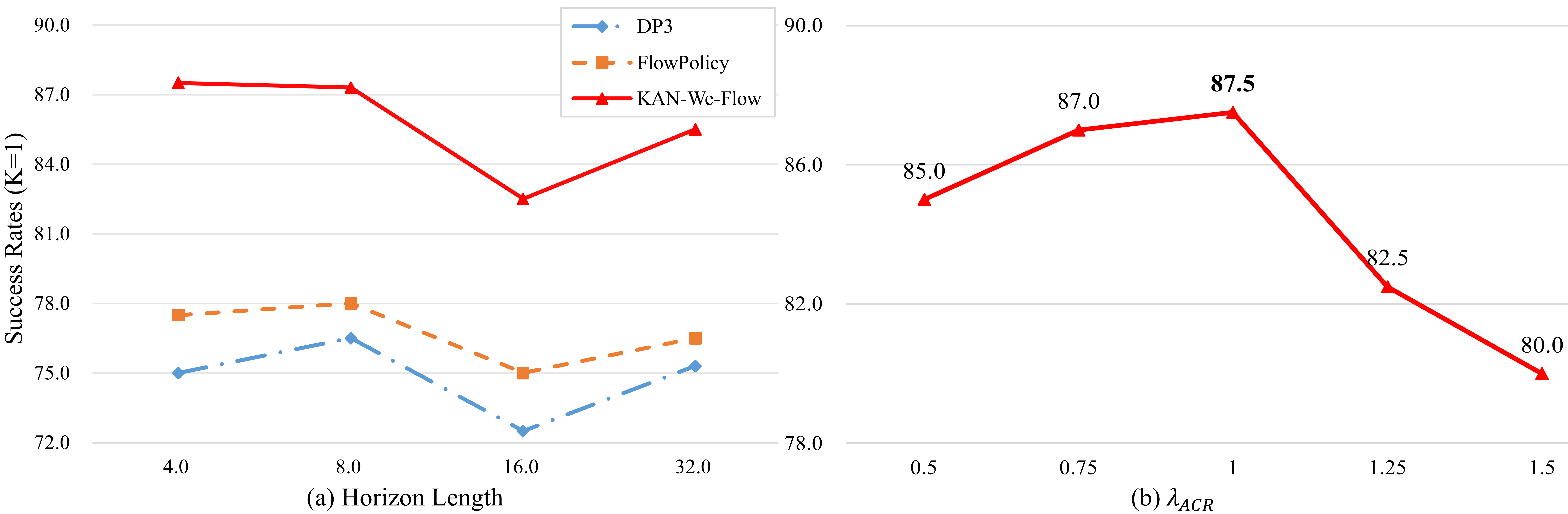} 
\caption{Ablations on horizon and hyperparameters. \textbf{(a)} Varying the prediction horizon: \emph{KAN–We–Flow} keeps stable and consistently outperforms DP3 and FlowPolicy. \textbf{(b)} $\lambda_{\text{ACR}}$ weight: performance peaks at $\lambda_{\text{ACR}}{=}1$; smaller values under-constrain end-point alignment, larger ones mildly over-regularize. } 
\label{fig:abla} 
\end{figure}
\subsection{Ablation Study}
\subsubsection{Ablations on Key Components}
We first assess the contribution of each proposed component on Adroit. Starting from the Baseline, progressively inserting RWKV time-mixing, RWKV channel-mixing, GroupKAN, and the Action Consistency Regularization (ACR) yields steady improvements in $SR1/SR3/SR5$ (Table~\ref{tab:table4}). Concretely, adding RWKV’s time-mixing already improves robustness, and further introducing the channel-mixing branch strengthens long-range temporal propagation; replacing the heavy per-channel MLP with GroupKAN consistently lifts all top-$K$ operating points by enabling compact, spline-based functional calibration; finally, the auxiliary ACR term provides the last increment by explicitly anchoring the one-step decode to expert behavior without extra inference steps. The full model (RWKV~+~GroupKAN~+~ACR) attains the best performance, substantiating the complementary roles of (i) linear-time sequence mixing (RWKV), (ii) parameter-efficient per-channel calibration (GroupKAN), and (iii) horizon-end anchoring (ACR).

\subsubsection{Ablation of Architectural Components within the RWKV–KAN Block}
We next replace the two core submodules in our RWKV-KAN block with strong alternatives to isolate where the gains come from (Table~\ref{tab:door}, Adroit–Door). Swapping RWKV for a CNN or an MHSA (Transformer) block improves over the baseline but remains notably behind our design, indicating that linear-time receptance-weighted mixing is particularly well-suited to action-trajectory modelling under tight latency budgets. Replacing RWKV with state-space variants (Mamba-V1/V2) yields larger gains than CNN/MHSA and narrows the gap, yet our RWKV block retains the top scores in $SR1/SR3/SR5$, suggesting that the receptance-gated, bidirectional scan we adopt better preserves long-horizon consistency in this setting. On the calibration side, substituting GroupKAN with a plain KAN or with an MLP confirms that learnable spline functions help, but the groupwise formulation with channel-affinity modulation is consistently stronger. 

\subsection{Ablations on Horizon Lengths and Hyperparameters}
Finally, we vary the action-prediction horizon to probe long-term dependency modelling (Fig.~\ref{fig:abla} a). Across short-to-long horizons, KAN–We–Flow maintains stable $SR1/SR3/SR5$ and outperforms DP3~\cite{dp3} and FlowPolicy~\cite{zhang2025flowpolicy} at every setting, indicating that (i) RWKV’s linear-time temporal mixing effectively propagates context over extended windows and (ii) ACR’s end-point anchoring mitigates exposure bias as the decode spans longer segments. The flat performance–horizon curve further suggests that our design achieves the desired efficiency–accuracy balance: it scales to longer horizons without resorting to multi-step refinement, while retaining real-time, one-step inference.

We ablate the hyperparameter $\lambda_{\text{ACR}}$ over a broad range while keeping other settings fixed. As shown in Fig.~\ref{fig:abla} b, performance peaks at $\lambda_{\text{ACR}}{=}1$; smaller values under-penalise end-point drift and larger values slightly over-regularise, with a relatively flat neighbourhood around the optimum indicating robustness.

\section{Conclusion}
In this work, we introduced KAN-We-Flow, a lightweight flow-matching visuomotor policy built on a synergistic RWKV–KAN UNet with a simple Action Consistency Regularization (ACR) term. Across Adroit, DexArt, and Meta-World, our model achieved state-of-the-art success rates while delivering practical efficiency: an 86.8\% parameter reduction relative to UNet-style diffusion baselines and single-step inference in the ~8–11 ms range, enabling ~100 Hz control. Ablations confirm that (i) RWKV’s linear-time temporal mixing and (ii) GroupKAN’s parameter-efficient, spline-based per-channel calibration are complementary, and that (iii) ACR stabilises one-step decoding, particularly for longer horizons. We believe these results demonstrate a clear path toward accurate, real-time, edge-deployable robot policies without heavy UNets. Future work will extend our approach to real-robot evaluations, richer multi-view sensing and language conditioning, and contact-rich, long-horizon tasks where safety and sim-to-real robustness are paramount.

\
\bibliography{icra}

\end{document}